\documentclass[preprint,times,sort&compress]{elsarticle}

\pdfoutput=1

\journal{ArXiv}

\usepackage{fixltx2e}

\usepackage{url}
\usepackage[breaklinks, hidelinks]{hyperref}

\usepackage{graphicx}

\usepackage{algorithmic}
\usepackage{algorithm}

\usepackage{amsmath}
\usepackage{mathtools}

\usepackage{multirow}
\usepackage{booktabs}
\usepackage{makecell}

\usepackage[format=hang, labelformat=parens]{subcaption}
\usepackage[acronym,smallcaps]{glossaries}
\newacronym{tfm}{TFM}{Transfer function of the impedance}
\newacronym{icc}{ICC}{Continuity of instrumentation}
\newacronym{orc}{ORC}{Ohmic resistance}
\newacronym{ls1}{LS1}{Long Shutdown 1}
\newacronym{hvq}{HVQ}{High Voltage Qualification}
\newacronym{doc}{DOC}{Dipole Orbit Corrector}
\newacronym{floss}{FLOSS}{Free/Libre and Open Source Software}	
\newacronym{www}{WWW}{World Wide Web}
\newacronym{ict}{ICT}{Information Communication Technologies}
\newacronym{loket}{LOKeT}{Building food market communities with the open source}	
\newacronym{te-mpe-ee}{TE-MPE-EE}{Technology Department, Machine Protection and Electrical Integrity group, Electrical Engineering}
\newacronym{fsf}{FSF}{Free Software Foundation}
\newacronym{mit}{MIT}{Massachusetts Institute of Technology}
\newacronym{cern}{CERN}{European Organization for Nuclear Research}
\newacronym{gpl}{GPL}{General Public Licence}
\newacronym{gplv2}{GPLv2}{GNU General Public License Version 2}
\newacronym{gplv3}{GPLv3}{GNU General Public License Version 3}
\newacronym{lgpl}{LGPL}{GNU Lesser General Public Licence}
\newacronym{osi}{OSI}{Open Source Initiative}
\newacronym{cpl}{CPL}{Common Public License}
\newacronym{epl}{EPL}{Eclipse Public License}
\newacronym{bsd}{BSD}{Berkeley Software Distribution}
\newacronym{agpl}{AGPL}{GNU Affero General Public License}
\newacronym{mpl}{MPL}{Mozilla Public Licence}
\newacronym{bsdnew}{BSD New}{BSD 3-Clause License}
\newacronym{freebsd}{FreeBSD}{BSD 2-Clause License}
\newacronym{mitlicence}{MIT}{MIT licence}
\newacronym{apachelicence}{Apache licence}{Apache License 2.0}
\newacronym{imap}{IMAP}{Internet Message Access Protocol}
\newacronym{pop}{POP}{Post Office Protocol} 
\newacronym{html}{HTML}{HyperText Markup Language}
\newacronym{rss}{RSS}{Rich Site Summary}
\newacronym{odt}{ODT}{Open Document Format}
\newacronym{lamp}{LAMP}{Linux, Apache, MySQL and PHP or Perl or Python}
\newacronym{vcs}{VCS}{Version Control System}
\newacronym{cvcss}{CVCSs}{Centralized Version Control Systems}
\newacronym{dvcs}{DVCs}{Distributed Version Control Systems}
\newacronym{lhc}{LHC}{Large Hadron Collider}
\newacronym{elqa}{ELQA}{Electrical Quality Assurance}
\newacronym{mscas}{MSCAs}{Mobile Social Computing Applications}
\newacronym{rca}{RCA}{Radio Corporation of America}
\newacronym{lep}{LEP}{The Large Electron-Positron Collider}
\newacronym{atlas}{ATLAS}{The Toroidal LHC Apparatus} 
\newacronym{cms}{CMS}{The Compact Muon Solenoid} 
\newacronym{alice}{ALICE}{A Large Ion Collider Experiment}
\newacronym{ssc}{SSC}{Superconducting Super Collider}
\newacronym{totem}{TOTEM}{The Total Cross Section, Elastic Scattering Diffraction Dissociation}
\newacronym{lugos}{LUGOS}{The Linux User Group of Slovenia}
\newacronym{xp}{XP}{Extreme Programming}

\makeglossaries

\begin{document}

\begin{frontmatter}

\title{A Conceptual Development of Quench Prediction App build on LSTM and ELQA framework}

\author[cernaddress]{Matej Mertik}
\ead{matej.mertik@cern.ch}

\author[aghaddress]{Maciej Wielgosz}
\ead{wielgosz@agh.edu.pl}

\author[aghaddress2]{Andrzej Skocze\'n}
\ead{skoczen@ftj.agh.edu.pl}

\address[cernaddress]{The European Organization for Nuclear Research - CERN, CH-1211 Geneva 23 Switzerland}
\address[aghaddress]{Faculty of Computer Science, Electronics and Telecommunications,\\AGH University of Science and Technology, Krak\'ow, Poland}
\address[aghaddress2]{Faculty of Physics and Applied Computer Science, AGH University of Science and Technology,\\Krak\'ow, Poland}

\begin{abstract}
This article presents a development of web application for quench prediction in  \gls{te-mpe-ee} at CERN. The authors describe an ELectrical Quality Assurance (ELQA) framework, a platform which was designed for rapid development of web integrated data analysis applications for different analysis needed during the hardware commissioning of the Large Hadron Collider (LHC). 
In second part the article describes a research carried out with the data collected from Quench Detection System by means of using an LSTM recurrent neural network. The article discusses and presents a conceptual work of implementing quench prediction application for \gls{te-mpe-ee} based on the ELQA and quench prediction algorithm.
\end{abstract}


\end{frontmatter}

\section{INTRODUCTION}
The Large Hadron Collider (LHC) is the world's largest and complex experimental facility ever created. It was built by the European Organization for Nuclear Research (CERN) between 1998 and 2008 in collaboration with over 10,000 scientists and engineers from over 100 countries, as well as hundreds of universities and laboratories. Its 27 kilometre circumference incorporates (among others) 1232 superconducting dipole magnets that lie beneath France -- Switzerland border near Geneva.

Starting up this largest and complex experimental facility requires many teams of professionals and complex procedures due to its superconducting characteristic in the domain of cryogenics, electrical engineering, electronics, computer science, material science, physics and others. 
The \gls{te-mpe-ee} at CERN provides an important task by hardware commissioning of the LHC machine, where all circuits are verified with many electrical measurements in the machine in different conditions during the complex procedure. 
The signals from measurements are stored in different systems at CERN, where data are available for their analysis. ELQA team started to develop a data analytic application to look through patterns implemented in the hardware commissioning in 2008 and 2014. As such a framework for prototyping data analysis apps was developed in the \gls{te-mpe-ee}. 

The ELQA framework provides building blocks for designing web application for data analysis. In this paper we present a conceptual development of Quench Prediction Application from the research work on the prediction of quenches and its preliminary results. 
We show the concept of integration of
\begin{enumerate}[(a)]
\item a research on a deep learning algorithm for quench prediction and
\item a research in software design for prototyping data analysis application for \gls{te-mpe-ee}.
\end{enumerate}
The paper is structured as following. 
In first chapters there is a description of the ELQA framework and software libraries used by software design, then LSTM concept for predicting quenches with deep learning is described on the real data. In the last part we are presenting a way and a concept of implementing the Quench Prediction App for \gls{te-mpe-ee} at CERN.

\section{ELQA FRAMEWORK}

\begin{center}
\begin{figure}
\centering
  \includegraphics[width=0.75\textwidth]{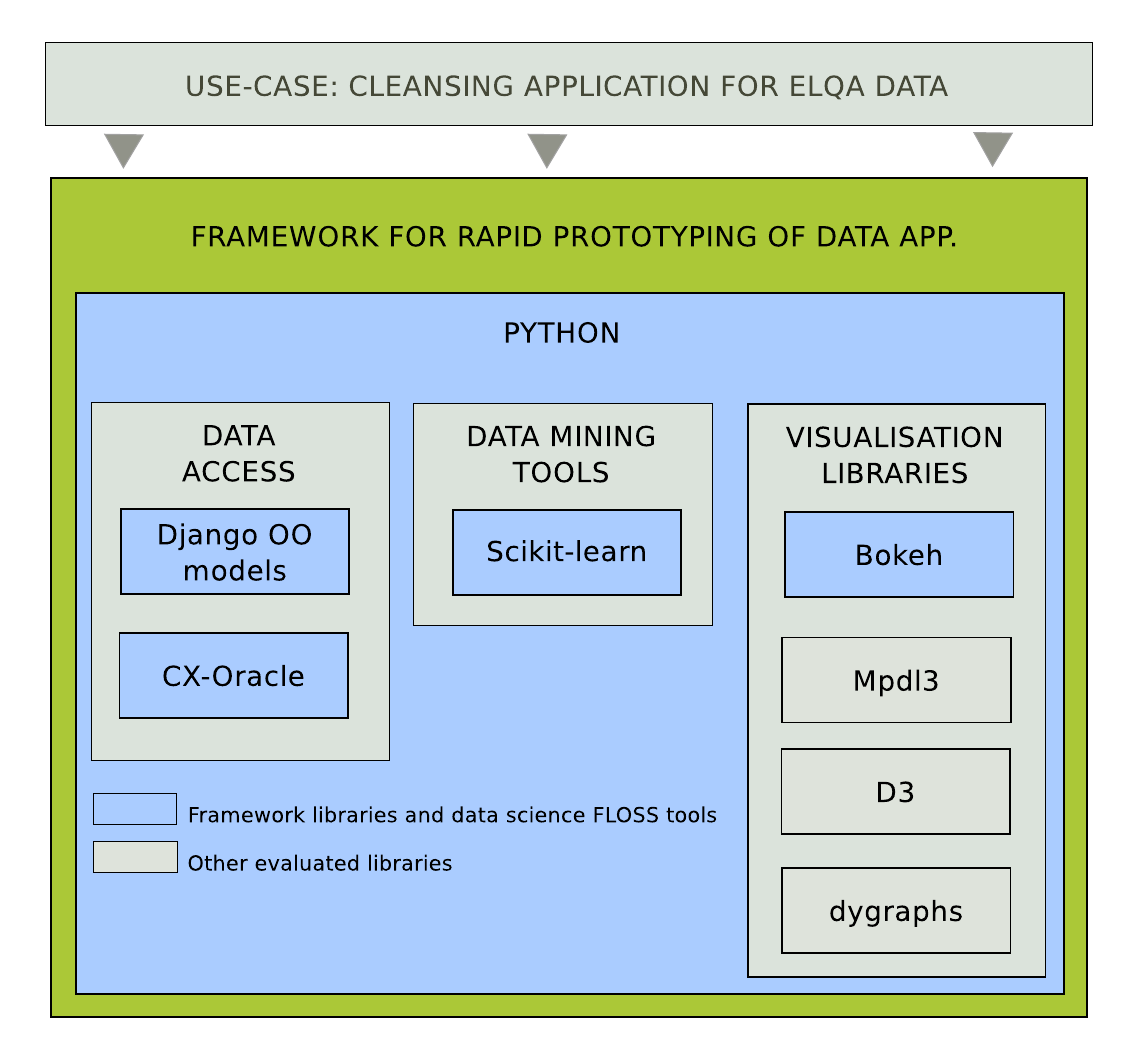}
\caption{The components of the framework for rapid design of data applications}
\label{fig_1}       
\end{figure}
\end{center}

ELQA framework was developed for rapid design of data analysis application at \gls{te-mpe-ee}. Framework was designed for \gls{elqa} group where there is a need to address a vast amount of data that are collected in the hardware commissioning of the \gls{lhc}. 
In the commissioning procedure the conditions of the circuits of the machine are verified and implemented by different measurements through all 8 sectors of the machine. Based on software design following requirements were identified carefully:
\begin{enumerate}[(a)]
\item searching signal's similarities over the vast amount of tests, 
\item visualizing trends of measurements through time domain in data, 
\item developing corresponding data models for different flavors of \gls{elqa} measurements such as TP4 (Test Procedure 4), DOC (Dipole Orbit Correctors), MIC (Magnet Instrumentation Check), 
\item developing different scenarios of visualization of different group of measurements.
\end{enumerate}
Two additional software characteristics were specially addressed:
\begin{enumerate}[(a)]
\item application should run in web environment and
\item framework should support high interaction for the user when exploring the data.
\end{enumerate}

As such ELQA framework provide different components for building the web applications. There are three main libraries used based on which the concept was defined:
\begin{enumerate}[(a)]
\item Bokeh Python interactive visualization library that targets modern web browsers for presentation \cite{Bokeh}, library was selected when testing the visualisation libraries and requirements \cite{lukematej},
\item Django high-level Python Web framework for Object Oriented paradigm to accessing relational dataset of \gls{elqa} \cite{Django}, and
\item Scikit-learn kit, a Machine Learning library used for the implementation of the KDD algorithms \cite{scikit-learn}.
\end{enumerate}
Within these three main libraries other dependent libraries and components of CERN software infrastructure were integrated. The relationships between them are shown in Fig. \ref{fig_1}.

Following subsections explain some detailed characteristic of the framework.

\subsection{ELQA FRAMEWORK CHARACTERISTICS}

\subsubsection{Data models}
Access to the data is addressed with the Object-Relational Mapping. Django framework handles this mapping providing all the functionality of a Structure Query Language (SQL). The architecture of Django is layered with
\begin{enumerate}[(a)]
\item the bottom layer which is the ELQA database, followed by
\item access library that is responsible for communication between Python and the database through SQL statements, here we used a Python library cx\_oracle \cite{cxOracle} and
\item specific Django database back-end where any kind of database supported by Django (PostgreSQL, MySql, SQLite and Oracle) is supported \cite{Django}. 
\end{enumerate}

Above Django's layers a data model is then defined for accessing the data of ELQA. Data models can be efficient programmed using the Django tools such as inspectdb for automatically defining the models from database tables, where relationships between the tables are separately defined.

\subsubsection{Preprocessors}
Preprocessors are classes for analytical preprocessing of the signals within defined operators by the ELQA engineers. For the extraction of signals currently following basic statistics are used as extractors (and can be extended): average, minimal value, maximal value, skewness, kurtosis and properties of the linear regression applied to each signal such as slope and standard error. Scipy and numpy libraries \cite{Scipy, Numpy} are used by preprocessors implementation.

\subsubsection{Data-miners}
Scikit-learn, a Machine Learning library written in Python is selected for data miners in the ELQA framework. It is built on NumPy, SciPy, and matplotlib and it is licenced under BSD open source licence \cite{scikit-learn}. Although there are many solutions and tools for data analysis available as open-source or commercial packages, some of the most used tools from the domains of Data Mining, Analytics, Big Data, and Data Science can be found on KDnuggets network maintained by a group of professionals and researchers \cite{KDnuggets}, Scikit-learn was selected as most appropriate as is written in Python and importantly has a strong community around the library.

The data-miner layer defines general class for analyzer. Currently two different clustering algorithms are integrated from the scikit-learn such as K-means clustering \cite{Kmeans} and DBScan clustering, a density-based spatial clustering of data with noise proposed by Martin Ester, Hans-Peter Kriegel, Jorg Sander and Xiaowei Xu in 1996 \cite{Ester96adensity-based}.

The analyzer class can be extended with other techniques from scikit-learn library or other libraries and algorithms in the framework. In the following section we present a concept of using LSTM-based solution for detecting and predicting quenches.

\subsubsection{Visualizers}
Visualizers serve user interaction and visualization of the data. When searching for appropriate tools in communities many frameworks in different programming languages were reconsidered. Due to the required architecture (see other evaluated libraries on the Fig. \ref{fig_1}) Bokeh was selected after studying the appropriate available solutions. Main reason was Bokeh's ability to act as a web server in order to expose a fully-fledged data visualization and analysis tool to multiple users \cite{lukematej}.

\section{CONCEPT OF USING THE LSTM-BASED SOLUTION}

The idea behind using LSTM algorithm for quench detection and prediction is based on unique and very useful properties of this deep learning algorithm in time series modelling. It turns out that in some cases a currently used quench protection solution \cite{qps_architecture, qps_reliablity} can be supplemented with an additional monitoring layer which monitors a performance of the Quench Protection System.

One of the biggest challenges in implementing of the LSTM concept is the construction of the network architecture by choosing a right set of hyper parameters such as number of layers, neurons within each layer and input size. This is done in trial-and-error fashion which can be substantially leveraged by using an appropriate framework. Such a framework allows for visualization of the performance a given network instance as well as easy manipulation of its macro parameter values. 

\subsection{RECURRENT NEURAL NETWORKS}
\label{section:rnn}

Phenomena occurring in a real world may be perceived in a spatial and temporal space. Since the neural models are supposed to mimic the real world the two different branches of neural networks evolved which address applications belonging to the aforementioned categories. Feed-forward and CNN networks are meant and most often used in applications of the spatial domain, whereas various kinds of recurrent models such as RNN, LSTM and GRU are especially useful in temporal modelling tasks. 

Unlike traditional models that are mostly based on hand-crafted features deep learning neural networks can operate directly on raw data. This makes them especially useful in applications where extracting features is very hard and even sometimes impossible. It turns out that there are many fields of applications where no experts exist who can handle feature extraction or the area is simply uncharted and we do not know whether the data contains latent patterns worth exploring \cite{Chang_2016_Application, Cai_2013_Deep, Toth_2013_Phone}. 

Foundations of the most currently used neural networks architectures were laid between 1960 and 1990. For almost the last two decades researchers were not able to take full advantage of these powerful models and there were claims that they are inherently ineffective. The whole machine learning landscape changed in early 2010 where deep learning algorithms started to achieve state-of-the-art results in a wide range of learning tasks. The breakthrough was brought about by several factors, among which computing power, deluge of widely available data and affordable storage are considered to be the critical ones. It is worth noting that in the presence of large amount of data the conventional simple linear models tend to under-fit or under-utilize computing resources.

\begin{figure}
\centering
\includegraphics[width=0.3\hsize]{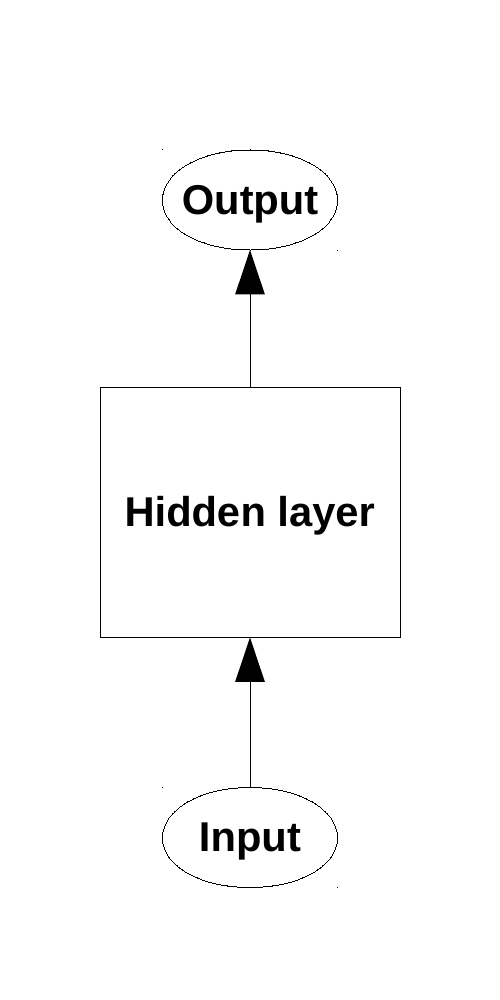}
\caption{Architecture of standard feed-forward neural network}
\label{fig:feedforward_overall_diagram}
\end{figure}

\begin{figure}
\centering
\includegraphics[width=1\hsize]{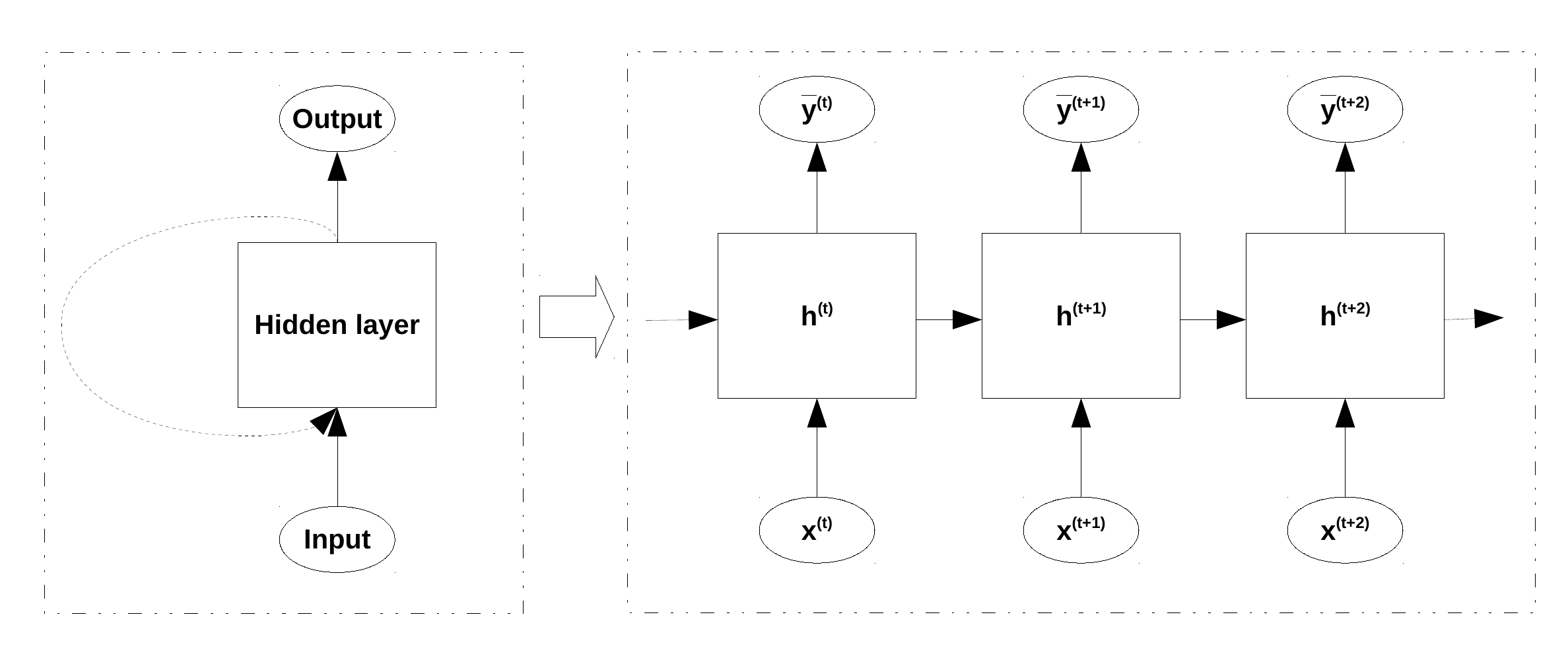}
\caption{General overview of recurrent neural networks}
\label{fig:rnn_overall_diagram}
\end{figure}

Deep learning models can operate in both spatial and temporal domain. However, standard feed-forward and CNN neural networks are well suited for spatial domain \cite{leCun_deep_2015, Farabet_Learning_2013, LeCun_deep_learning_2015}, in contrast to recurrent neural networks which perform much better in the applications of temporal domain \cite{Morton_2016_Analysis, Pouladi_2015_Recurrent, Chen_2016_Efficient}. CNNs and feed-forward networks rely on the assumption of independence of data within training and testing set as presented in Fig. \ref{fig:feedforward_overall_diagram}. This means that after each training item is presented to the model, the current state of the network is lost i.e. temporal information is not taken into account in training a model. 

In a case of the independent data it is not an issue. But for data which contain crucial time or space relationships it may lead to a loss of majority of the information which is located in between steps. Additionally, feed-forward models expect fixed length of training vectors which is not always the case, especially when dealing with time domain data.

Recurrent neural networks (RNNs) are models with the ability to process sequential data as one element at a time. Thus they can simultaneously model sequential and time dependencies on multiple scales. Unfortunately, a range of practical applications of standard RNN architectures is quite limited. This is caused by the influence of a given input on a hidden and output layer during the training of the network. It either decays or blows up exponentially as it moves across recurrent connections. This effect is described as the \textit{vanishing gradient problem} \cite{bookAGraves}. 
There were many unsuccessful attempts to address the problem before LSTM was eventually introduced by Hochreiter and Schmidhuber \cite{Hochreiter_Long_1997} which ultimately solved it.

Recurrent neural networks may be visualized both as a looped-back architectures of interconnected neurons which was presented in Fig. \ref{fig:feedforward_overall_diagram}. Originally RNNs were meant to be used with single variable signals but they were also adapted for multiple stream inputs \cite{bookAGraves}.

\begin{figure}
\centering
\includegraphics[width=0.3\hsize]{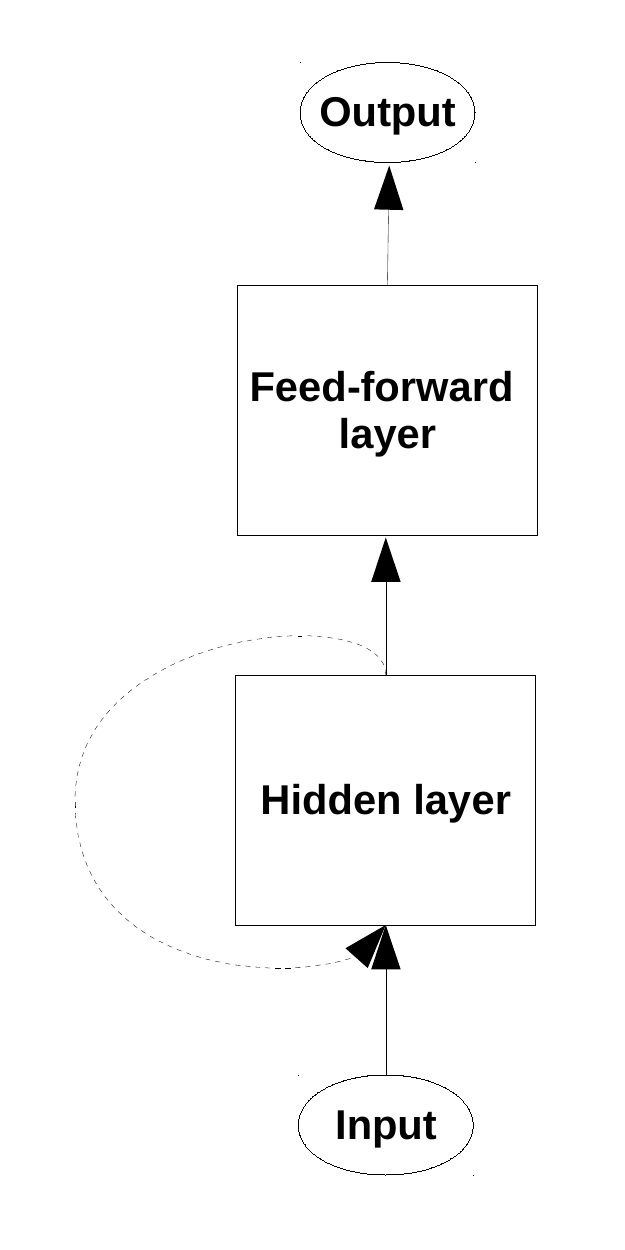}
\caption{Mapping RNN outputs.}
\label{fig:ffn_over_rnn_diagram}
\end{figure}

\begin{figure}
\centering
\includegraphics[width=0.9\hsize]{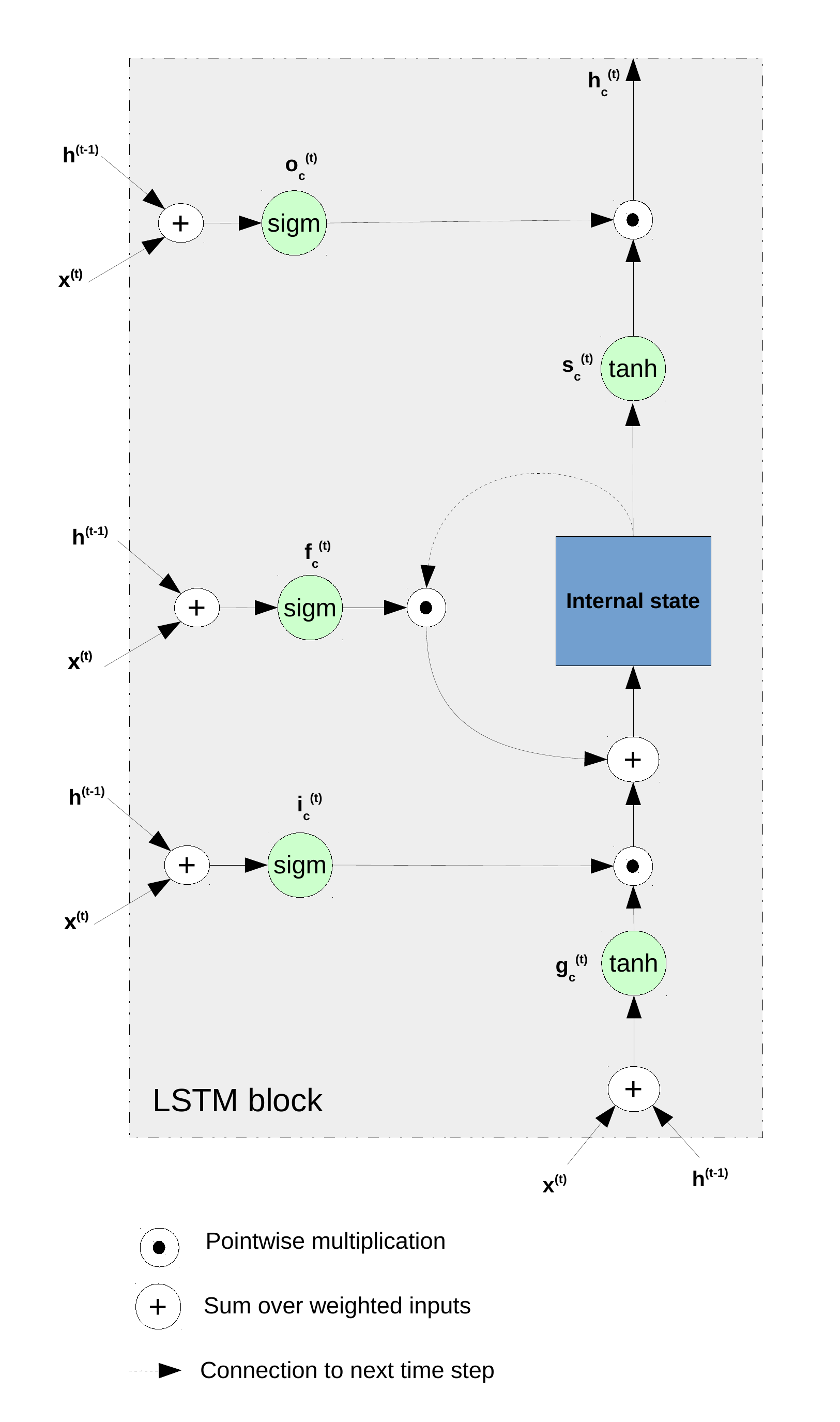}
\caption{Architecture of LSTM memory block containing one memory cell}
\label{fig:lstm_block_architecture}
\end{figure}

It is a common practice to use feed-forward recurrent layers together in order to map outputs from RNN or LSTM to the result space as presented in Fig. \ref{fig:ffn_over_rnn_diagram}. 

In practical applications, the LSTM has shown extraordinary ability to learn long-range dependencies as compared to standard RNNs. Therefore, most of the state-of-the-art applications use LSTM model \cite{Zachary_2015_Critical}. 

The LSTM internal structure is based on a set of connected memory blocks as presented in Fig. \ref{fig:lstm_block_architecture}. The memory blocks contain \textit{memory cells} with loop-backed connections storing the temporal state of the network in addition to boundary units called \textit{cell} which serve as an interface for information propagation within the network. 

There are three different gates in each memory block:
\begin{enumerate}[(a)]
 \item $input$ gate which controls input activations into the memory cell,
 \item $output$ gate controls cell outflow of activations into the rest of the network,
 \item $forgate$ gate scales the internal state of the cell before summing it with the input through the self-recurrent connection of the cell. This enables gradual forgetting of a cell memory state.
\end{enumerate}

Modern LSTM architectures may also contain \textit{peephole connections} \cite{Greff_2015_LSTM}. Since we do not use in our experiment it was not depicted in Fig. \ref{fig:lstm_block_architecture} and addressed in this description.

The output of each LSTM block is calculated according to the following set of equations:

\begin{equation}
g^{(t)} = \phi(W_{gx}x^{(t)} + W_{gh}h^{(t-1)} + b_g)
 \label{eq:lstm_input_node}
\end{equation}

\begin{equation}
i^{(t)} = \sigma(W_{ix}x^{(t)} + W_{ih}h^{(t-1)} + b_i)
 \label{eq:lstm_input_gate}
\end{equation}

\begin{equation}
f^{(t)} = \sigma(W_{fx}x^{(t)} + W_{fh}h^{(t-1)} + b_f)
 \label{eq:lstm_forget_gate}
\end{equation}

\begin{equation}
o^{(t)} = \sigma(W_{ox}x^{(t)} + W_{oh}h^{(t-1)} + b_o)
 \label{eq:lstm_output_gate}
\end{equation}

\begin{equation}
s^{(t)} = g^{(t)} \odot i^{(t)} + s^{(t-1)} \odot f^{(t)}
 \label{eq:lstm_internal_state}
\end{equation}

\begin{equation}
h^{(t)} = \phi (s^{(t)}) \odot o^{(t)}
 \label{eq:lstm_internal_state_phi}
\end{equation}

While examining equations \ref{eq:lstm_input_node}, \ref{eq:lstm_input_gate}, \ref{eq:lstm_forget_gate}, \ref{eq:lstm_output_gate} and \ref{eq:lstm_internal_state} it may be noticed that instances for a current and previous time step are used for the value of hidden layer vector $h$ as well as for internal state $s$. Consequently, $h^{t}$ denotes a value of a hidden state vector at a current time step, whereas $h^{t-1}$ refers to previous step. It is also worth noting the equations contain vector notation which means that they address whole set of LSTM cells. In order to address a single cell a subscript $c$ is used as it is presented in Fig. \ref{fig:lstm_block_architecture} where for instance $h_c^{t}$ refers to the value of hidden state within this particular cell.

LSTM based network learns when to let activation into the internal states of its cells and when to let the activation out. This is implemented with gating mechanism in which all the gates are considered as separate constitutes of LSTM block with their own learning capability. This means that they adapt in a training process as separate units to preserve a proper information flow. When both the gates are closed, the activation is cut-off from the outside connections, neither growing or shrinking, nor affecting the output at intermediate time steps. In order to make it possible a hard sigmoid function $\sigma$ was used which can output 0 and 1 as given by Eq. \ref{eq:hard_sigma}. This means that the gates can be fully open or closed.

\begin{equation}
\begin{split}
\sigma (x) =
\begin{cases}  
0 \text{~if~}  x \leq t_{l} \\ 
ax + b  \text{~if~} x \in (t_{l}, t_{h}) \\
1 \text{~if~}  x \geq t_{h} \\ 
\end{cases}
\label{eq:hard_sigma}
\end{split}
\end{equation}

In terms of the backward pass, so-called \textit{constant error carousel} enables the gradient to propagate back through many time steps, neither exploding or vanishing \cite{Hochreiter_Long_1997, Zachary_2015_Critical}.

\subsection{EXPERIMENTAL SETUP}
\label{subsection:experiments_setup}

Timber database stores a record of many years of the magnets activity. This is a huge amount of data with relatively few quench events. The first research question we had to address was related to a time span surrounding a quench event which should be taken into account for the LSTM model training. One day-long record of a single voltage time series for only one magnet occupies roughly 100 MB and there are several voltage time series associated with a single magnet \cite{timber} but ultimately we decided to use only one in our experiments, namely $U_{res}$. 

Various kinds of magnets are used in LHC and they generate different number of quench events. It is highly beneficial from a perspective of our research to pick a magnet for which large possible number of quenches is provided. 
The longest history of quenches was provided for 600A magnets in Timber database. Therefore, we decided to focus our initial research on 600A magnets. Unfortunately, the data covering a quench periods of 600A magnets is very large i.e. an order of several gigabytes. However, as it was mentioned before, the activity record of superconducting magnets during operational time of LHC is composed mostly of sections of normal work and partially of quench events. Furthermore, Timber does not enable automated quench periods extraction, despite having many useful features for data preprocessing and information extraction. 

Training deep learning models takes a long time even when fast GPUs are employed. Therefore, a choice of a size of a training data and a set of hyper-parameters of a model is critical in terms of the application performance. In course of experiments we discovered that it is faster to first train a model with a small data set and preliminarily adjust hyper-parameters and then tweak them on bigger data sets.

Consequently, we have created 3 different data sets: small, medium and the large ones as presented in Tab. \ref{tab:data_sets}.

\begin{table}
\caption{Data sets used for training and testing the model}
\label{tab:data_sets}
\centering
\begin{tabular}{lc}
\toprule
Data set & size [MB] \\ 
\midrule
small & 22  \\
medium & 111 \\
large & 5000  \\
\bottomrule
\end{tabular}
\end{table}

The tests are in progress with a main goal of validating the LSTM-based setup capability of modeling superconducting magnets behavior. The resolution of the data acquired from Timber database is far too low to determine its ability to predict or detect quenches. This is will require using more data of higher resolution, such as post-mortem records, and is planned as the future work.

\section{INTEGRATING LSTM MODEL INTO ELQA FRAMEWORK}

Described LSTM concept developed at TE-MPE-EE can be efficiently used within ELQA framework for implementing quench prediction application for the end user. Following characteristics and key benefits are identified for joining these two pieces of research
\begin{enumerate}[(a)]
\item efficient application of the research conducted on quench detection and prediction,
\item incorporating quench detection application into web environment
\item need of an efficient graphical user interface for end user,
\item availability of high user interaction experience on the developed application for quench detection.
\end{enumerate}

\begin{center}
\begin{figure}
  \includegraphics[width=1.0\textwidth]{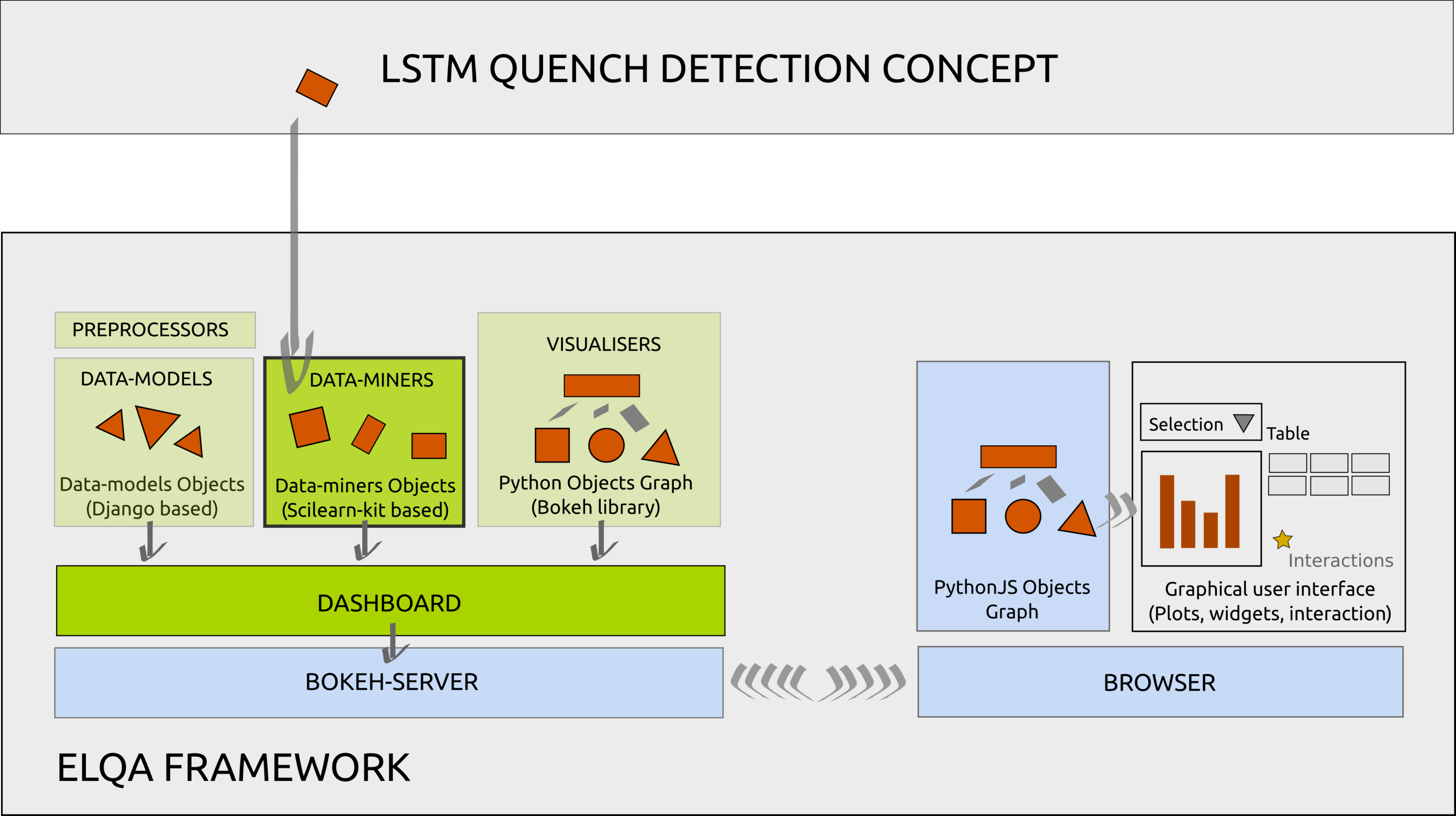}
\caption{Integration of framework and LSTM}
\label{fig_2}       
\end{figure}
\end{center}

Fig. \ref{fig_2} presents possible integration of the LSTM into the framework. It can be seen that LSTM concept should be wrapped by the new LSTM analyzer class and then used in the ELQA framework. The graphical user interface is a next step that should be designed through the definition of the Quench Prediction App Dashboard, a key part for any web apps developed for data analysis based on ELQA.

Following subsection provides an overview for integration process of the deep learning LSTM.

\subsection{INTEGRATING LSTM-BASED SOLUTION}

\begin{center}
\begin{figure}
  \includegraphics[width=0.9\textwidth]{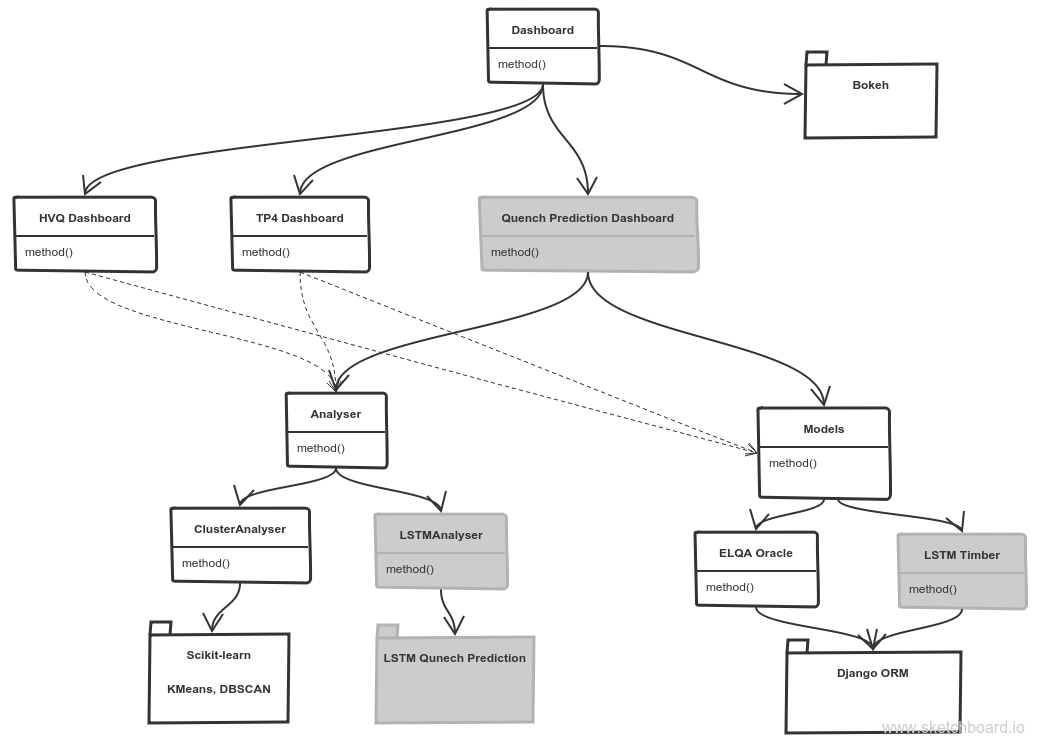}
\caption{Integration of framework and LSTM, grey entities presents LSTM quench prediction parts in the ELQA framework}
\label{fig_integration}       
\end{figure}
\end{center}

Analyser class in ELQA is a general class where data miner algorithms are integrated from Scikit-learn data mining library. LSTM concept for quench prediction is implemented in Python at \gls{te-mpe-ee} and can be logical integrated into ELQA with an extended LSTM from the Analyser.

Accessing the data in Timber database will be extended by new Data model. ELQA framework allows to define new stack for integrating the access to the various datasets and stacks through Django library. Fig. \ref{fig_integration} presents a possible integration.

When model and algorithm are incorporated into the ELQA classes, full development of the dashboard can be then applied through the framework for user interface and interaction developed for a Quench prediction app.

\subsection{IMPLEMENTING QUENCH PREDICTION DASHBOARD - APP DEVELOPMENT}

\begin{center}
\begin{figure}
  \includegraphics[width=0.9\textwidth]{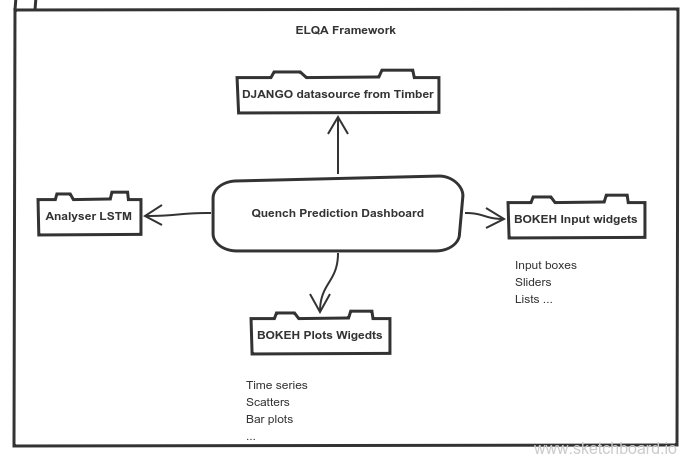}
\caption{Dashboard design with ELQA for Quench Prediction app}
\label{fig_dashboard}       
\end{figure}
\end{center}

With incorporated data model to access to Timber dataset and wrapped LSTM algorithm of Quench Prediction Software within Analyser class, efficient development of the web application for Quench prediction can be applied through programming of inherited LSTM dashboard. IN LSTM Dashboard widgets for Graphical User Interface are defined and then integrated through Bokeh library (input forms, plots), where data are applied to LSTM algorithm. Fig. \ref{fig_dashboard} presents the concept of the Dashboard implementation by ELQA.

It can be seen (Fig. \ref{fig_dashboard}) that conceptual building blocks when programming Dashboard for quench prediction application follow objects that need to be defined in the Dashboard:
\begin{enumerate}[(a)]
\item input widget for user interaction (lists, input boxes, sliders etc..),
\item plots widget for plotting series,
\item data sources from the data model (Django model for accessing Timber) and 
\item LSTM algorithm as data-miner for quench prediction application.
\end{enumerate}

\section{CONCLUSIONS}
We presented a concept paper of integrating the LSTM techniques for quench detection into the web application used the ELQA framework for prototyping data analysis application developed at \gls{te-mpe-ee} at CERN. Main characteristics of the framework were presented as the concept of LSTM for quench detection. We showed the mechanism and key benefits of joining these two entities for developing a quench prediction app for TE-MPE-EE at CERN. As explained, the LSTM concept can be integrated efficiently in order to provide an application that can be used by the engineers when providing hardware commissioning procedures of the LHC at CERN.

\section*{Acknowledgments}
The framework is being tested to prototype data application by the ELQA team. We would like to thank the entire ELQA team and TE-MPE-EE at CERN for their persistent invaluable help throughout the development process.

\bibliographystyle{elsarticle-num}
\bibliography{references}

\end{document}